%% file: Gdropout.tex
\documentclass[journal]{IEEEtran}

\input{preamble}
\begin{document}
\title{Data Dropout in Arbitrary Basis for Deep Network Regularization}

\author{Mostafa~Rahmani, \IEEEmembership{Student Member,~IEEE}, and George~K.~Atia,~\IEEEmembership{Member,~IEEE} 
\thanks{
This work was supported in part by NSF Award CCF-1552497.

The authors are with the Department of Electrical and Computer Engineering, University of Central Florida, Orlando, FL 32816 USA (e-mails: mostafa@knights.ucf.edu, george.atia@ucf.edu).}
}

\markboth{}%
{Shell \MakeLowercase{\textit{et al.}}: Bare Demo of IEEEtran.cls for Journals}
\maketitle

\begin{abstract}
An important problem in training deep networks with high capacity is to ensure that the trained network works well when presented with new inputs outside the training dataset. Dropout is an effective regularization technique to boost the network's generalization in which a random subset of the elements of the given data and the extracted features are set to zero during the training process.
In this paper, a new randomized regularization technique in which we withhold a random part of the data without necessarily turning off the neurons/data-elements is proposed.
In the proposed method, of which the conventional dropout is shown to be a special case, random data dropout is performed in an arbitrary basis, hence the designation Generalized Dropout. We also present a framework whereby the proposed technique can be applied efficiently to convolutional neural networks. The presented numerical experiments demonstrate that the proposed technique yields notable performance gain.
Generalized Dropout provides new insight into the idea of dropout, shows that we can achieve different performance gains by using different bases matrices, and opens up a new research question as of how to choose optimal bases matrices that achieve maximal performance gain.

\end{abstract}


\IEEEpeerreviewmaketitle

\section{Introduction}
In recent years, deep neural networks have made major breakthroughs in computer vision, natural language processing, and many other applications of machine learning \cite{zeiler2014visualizing,sutskever2014sequence,vinyals2015neural,cho2014learning,bahdanau2014neural,sermanet2013overfeat,he2016deep,Goodfellow-et-al-2016
}. Deep networks have remarkable capacity and can merge low and high-level features in an end-to-end multi-layer and differentiable framework.  The modern networks are highly non-linear and have millions of  parameters \cite{zagoruyko2016wide}. These features make them notably expressive and capable of learning complicated data models. However, when the training data is limited, overfitting may occur
in which the network could end up learning irrelevant patterns from the training data which can significantly degrade the generalizability of the learned model.
While overfitting is best alleviated by using more data for training, obtaining labeled data can be quite expensive in practice. Thus, network regularization is one of the central problems in machine learning and many ideas have been proposed to reduce overfitting. We briefly review some of the  deep network regularization techniques. Subsequently,  the proposed random data dropout method is presented and validated through numerical simulations. It is shown that the proposed regularization technique can yield a notable performance gain. The idea of data dropout does not necessarily require to randomly shut down the data-elements/network-neurons. The conventional data dropout method is in fact a special case of the proposed approach.

\subsection{Regularization in deep learning}
One of the early regularization techniques is to introduce norm penalties of the network weights (such as $\ell_1$-norm or $\ell_2$-norm  penalties) in the training cost function \cite{Goodfellow-et-al-2016}. These penalty functions limit the network's capacity because adding norm penalties is equivalent to constraining the network weights to be inside some norm balls. Thus, an equivalent method is to directly control the $\ell_2$-norm or $\ell_1$-norm of the weights during the training process \cite{hinton2012improving,srebro2005rank}.  A different idea is to enforce a desired structure to the weights. A successful example of this technique is the Convolutional Neutral Network (CNN) in which the spatial invariance is directly obtained by the convolution-based design of the weight matrices. Early stopping is another simple regularization technique which stops the learning process once the error on a separate validation set starts increasing. In other words, it seeks to prevent the network from becoming over-specific to the training data and learning irrelevant patterns. 

\begin{figure}[t!]
 \centering
    \includegraphics[width=0.5 \textwidth]{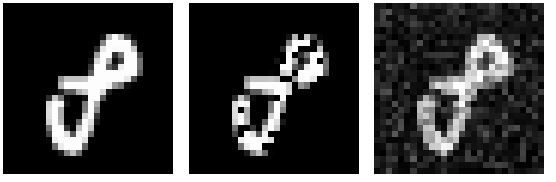}
    \vspace{-0.3cm}
    \caption{Left: An MNIST training image which depicts integer 8. Middle: The image shown on the left after setting each element to zero independently with probability 0.2. Right: The training image shown on the left after applying the projection matrix as in (\ref{eq_drop2}) with $p = 0.2$ (the basis used in this example is a randomly generated basis). }
    \label{fig:fake_data}
\end{figure}

This paper focuses on an effective regularization technique, which is to generate fake training data (data augmentation). In order to obtain valid fake data from a training example, a transformation is applied to the data such that it does not change its class. For instance, in object recognition, image translation, flipping, and rotation have proved to be effective to this end. Another data augmentation technique is to add perturbation noise to the  data \cite{poole2014analyzing}.
A different idea for generating fake data is by withholding part of the data given that an efficient network is expected to correctly classify the data from the reduced information. For instance, imagine an image of a horse showing the legs, the body, and the head. An intelligent classifier should be able to identify the horse by seeing only some part of the image such as the part that shows the head of the horse. Dropout uses a similar idea to generate artificial training data \cite{srivastava2014dropout}. It randomly sets to zero a random subset of the elements of a training example. For instance,  the middle image of Fig. \ref{fig:fake_data} shows the original image (on the left) after applying Dropout where each element is set to zero with probability equal to 0.2. Interestingly, the Euclidean distance between a data point and the same data point after dropout is large, but we expect the network to assign them the same class label. In recent years, many variants of the dropout method first introduced in \cite{srivastava2014dropout} were presented \cite{kingma2014adam,wager2013dropout,rennie2014annealed,li2016improved,bayer2013fast,wang2013fast,wu2015towards,kubo2016compacting,ba2013adaptive}. Given that the randomized technique proposed in \cite{srivastava2014dropout} is a special case of our method, all these techniques are applicable to our proposed approach.



\section{Proposed Approach}
Suppose that $\bx \in \mathbb{R}^{N \times 1}$ is a training example, where $N$ is the ambient dimension.
Define $\{ \be_i \}_{i=1}^N$ as the  standard basis, where $\be_i \in \mathbb{R}^{N \times 1}$ is a vector whose elements are all zero except for the $i^{\text{th}}$ element, which is equal to 1.
The training example $\bx$ can be written as
\begin{eqnarray}
\bx = \sum_{i=1}^{N} x_i \be_i  \:,
\end{eqnarray}
where $x_i = \be_i^T \bx$. If we define $\bx_d$ as the vector $\bx$ after applying the random data dropout, then $\bx_d$ can be represented as
\begin{eqnarray}
\bx = \sum_{i=1}^{N} \alpha_i x_i \be_i  \:,
\end{eqnarray}
where $\{\alpha_i \}_{i=1}^N$ are independent binary random variables with the probability $\mathbb{P}\{\alpha_i = 0\} = p$.
However, an $N$-dimensional space can be spanned by an infinite number of different bases.
For instance, suppose $\bG \in \mathbb{R}^{N \times N}$ is an orthonormal matrix different from the identity matrix\footnote{ For instance, we can build $\bG$ by sampling a random unit $\ell_2$-norm vector. Then, Matrix $\bG$ can be formed by concatenating the random vector and a orthonormal basis for the subspace which is orthogonal to the randomly chosen vector.}.
Thus, the vector $\bx$ can be represented with respect to the columns of $\bG$ as
\begin{eqnarray}
\bx_d = \sum_{i=1}^{N}  \left( \bg_i^T \bx \right)  \bg_i  \:.
\label{eq_drop1}
\end{eqnarray}
Accordingly, we propose to perform the random data dropout with respect to $\bG$ as
\begin{eqnarray}
\bx_d = \sum_{i=1}^{N} \alpha_i \left( \bg_i^T \bx \right)  \bg_i  \:.
\label{eq_drop1s}
\end{eqnarray}
If we define $$ \bP_d = \sum_{i}^N \alpha_i \bg_i \bg_i^T\:,$$
then (\ref{eq_drop1s}) is equivalent to
\begin{eqnarray}
\bx_d =\bP_d \bx \: .
\label{eq_drop2}
\end{eqnarray}
In each training iteration, we sample the binary random variables to form the matrix $\bP_d$.
In contrast to the conventional dropout, we can choose different basis matrices for the different layers of the network. In addition, we can change the basis matrices over the course of training. For instance, in the presented numerical simulations, we generate a new random basis after each epoch.
As an example, Algorithm 1 demonstrates the proposed approach, Generalized Dropout, in a forward pass of a $k$-layer fully connected network during the training process. In Algorithm 1, $\{\bW_i\}_{i=1}^k$ and $\{ \bb_i\}_{i=1}^k$ are defined as the weight matrices and bias vectors of the network, respectively. In addition, $N$ is defined as the dimension of the given data and $N_k$ as the dimension of the data after the $k^{\text{th}}$ layer.

\begin{algorithm}
\caption{Generalized Dropout in a $k$-layer fully connected network }

\textbf{Input:} Data example $\bx \in \mathbb{R}^{N}$.

\smallbreak

\textbf{Initialization:} Define $\{ \bG_i \}_{i=0}^k$ as orthonormal bases where $\bG_0 \in \mathbb{R}^{N \times N}$ and $\bG_i \in \mathbb{R}^{N_i \times N_i}$ for $1 \leq i \leq k$, respectively.

\smallbreak

\textbf{1. Apply the proposed data dropout to the given data:}

\textbf{1.1} Form $\bP_d = \sum_{j=1}^{N} \alpha_j \bg_j^0 \bg_j^0$, where $\{ \alpha_j \}_{j=1}^N$ are independent binary random variables and $\bg_j^0$ is the $j^{\text{th}}$ column of $\bG_0$.

\textbf{1.2} Update $\bx 	\gets \bP_d \bx \:. $

\smallbreak

\textbf{2. Apply the proposed data dropout to all the layers}\\
\textbf{2.1 For} $i$ from 1 to $k$

\textbf{2.1.1} Apply the weight matrix and bias vector: $\bx 	\gets \bW_i \bx + \bb_i$.

\textbf{2.1.2} Apply the element-wise non-linear function to $\bx$.

\textbf{2.1.3} Form $\bP_d = \sum_{j=1}^{N_i} \alpha_j \bg_j^i \bg_j^i$, where $\{ \alpha_j \}_{j=1}^{N_i}$ are independent binary random variables and $\bg_j^i$ is the $j^{\text{th}}$ column of $\bG_i$.

\textbf{2.1.4} Apply the proposed data dropout: $\bx 	\gets \bP_d \bx$.

\textbf{2.1 End For}

\smallbreak

\textbf{3. } Project $\bx$ to the network output.
\end{algorithm}

During the course of training, a neural network observes a training example several times. If the network is highly expressive, it can become over-specific to the training example. However, with the random data dropout, the network experiences different variants of a training example in different iterations. These randomly generated training data are relatively far apart in the Euclidean space. Thus, the random data dropout can help the network to not be over-specific to the training examples.
In a multi-layer neural network, the output of each layer plays the role of the data for the network composed of the next layers. Thus, we apply random data dropout to the features' vectors of a network layer to generate random fake data for the subsequent layers.

In the proposed approach, during testing we ignore the data projection steps and scale the data properly similar to the scaling suggested in \cite{srivastava2014dropout}.

\subsection{Application  to convolutional networks }
In a CNN, the data corresponding to each training example is a  third order tensor. Suppose $\bx \in \mathbb{R}^{c \times a \times b}$ is a data example after a convolution layer, where $c$ is the number of output channels, and $a$ and $b$ are the height and width of each plane, respectively.
In order to perform the proposed data dropout, we can apply the random projection in multiple ways. \\
The first way is to reshape the data into a vector such that $\bx \in \mathbb{R}^{ c \: a \: b}$ and apply the proposed data dropout to the vectorized $\bx$. Then, we can reshape the data back into the original third order tensor.

The second and a more computationally efficient way is to perform the data dropout along the feature planes dimension. In other words, imagine that the third order tensor $\bx \in \mathbb{R}^{c \times a \times b}$ is a vector whose elements are matrices of dimension $a\times b$. Accordingly, we can apply the random data dropout by applying the projection matrix as a linear operator to this vector. Suppose the dimension of the projection matrix $\bP_d$ is $c\times c$ and we have reshaped the third order tensor $\bx$ into a matrix of size $c \times (a\:b)$. Then, the data dropout is performed by applying the projection matrix to the reshaped data as $\bx \gets  \bP_d \bx$. After the data projection, we can reshape the data back to the third order tensor to prepare it for the next convolutional layer.

The third way is to apply the projection matrix to the feature planes. In other words, first $\bx \in \mathbb{R}^{c \times a \times b}$ is reshaped into $\bx \in \mathbb{R}^{(a\:b) \times c}$ then the data dropout is performed $\bx \gets \bP_d \bx$ in which $\bP_d \in \mathbb{R}^{(a\:b) \times (a\: b)}$. In addition, we can combine the second and third way. In the presented numerical experiments, we use the second way, i.e., the random projection is applied along the feature planes dimension ($\bP_d \in \mathbb{R}^{c\times c}$).

\section{Numerical Experiments}
In this section, we validate the proposed approach by applying Generalized Dropout to two neural networks. In addition, we compare the performance gains achieved by the proposed method with different basis matrices (including the identity matrix which results in the conventional dropout proposed in \cite{srivastava2014dropout}).
In the first experiment, a 3-layer  fully-connected network is implemented. In the second experiment, a 5-layer CNN is used. The element-wise  non-linear function used in all the experiments is the rectified linear unit (ReLU) \cite{jarrett2009best}. We use the MNIST data set which consists of 60000 training and 10000 test examples each representing a $28 \times 28$ digit image.
In all the experiments, we use three choices for the matrix $\bG$. One choice is the identity matrix $\bI$, which corresponds to the conventional dropout.
The second choice is a normalized Hadamard matrix. To be able to use the Hadamard matrix as the basis matrix, we zero pad the given image data changing its size from $28 \times 28$ to $32 \times 32$. Our third choice is a random orthonormal basis. To form an $N \times N$ random orthonormal basis, first we build an $N \times N$ matrix with independent Gaussian random variables and use its left singular vectors as the basis matrix (matrix $\bG$). 
In order to form the matrix $\bP_d$ for the input data, we set $p = 0.2$. For the next layers, $p = 0.5$.

\subsection{Fully-connected network}
\begin{figure}[t!]
 \centering
    \includegraphics[width=0.38\textwidth]{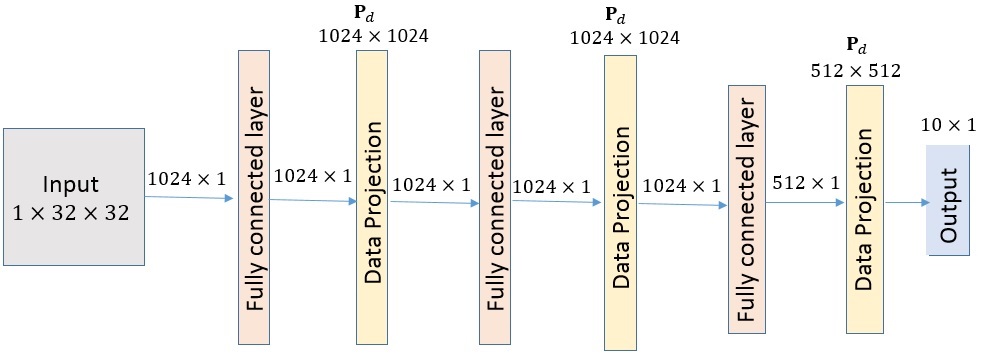}
    \vspace{-0.3cm}
    \caption{The structure of the fully-connected network (during training process) used in the first experiment.}
    \label{fig:fully_connected}
\end{figure}

\begin{figure}[t!]
 \centering
    \includegraphics[width=0.4 \textwidth]{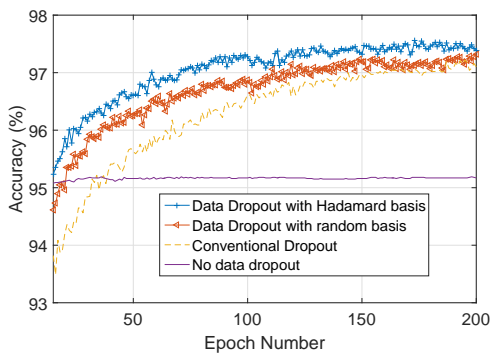}
    \vspace{-0.3cm}
    \caption{The test error versus epoch number for the fully connected network. The network is trained with 20000 images.}
    \label{fig:test_error_full_c}
\end{figure}

In this experiment, we apply Generalized Dropout to all the layers of a 3-layer fully-connected neural network.  Fig. \ref{fig:fully_connected} illustrates the network structure.
In this structure, we have 4 data dropout steps where the first is applied to the given data. Fig. \ref{fig:fully_connected} shows the dimension of the data after each layer of the network.
In order to accentuate the overfitting phenomena,  the network is not trained with the whole training  data. Instead, the data used for training in this example is 20000 images randomly sampled from the MNIST training data.
We use the cross-entropy cost function and  the network parameters are updated using stochastic gradient descent with a momentum term of 0.9 \cite{sutskever2013importance}.
Overfitting does not occur in the early stages of the training process. Thus, we first train the network for 50 epochs without any data dropout.  We train the saved network for 200 epochs using different basis matrices for random data dropout. In the first 120 epochs the learning rate is $10^{-3}$ and in the last 80 epochs the learning rate is reduced to $10^{-3}/3$. The size of mini-batches is 32.
Fig. \ref{fig:test_error_full_c} shows the test error of the network after each epoch.
One can observe that the best performance gains achieved by the Hadamard and the random bases are higher than the corresponding gain by the conventional dropout. Table \ref{tab:gain_f} summarizes the best performance gains. How to find the best basis matrix to achieve the maximal performance gain is an interesting question for future research.

\begin{table}
\centering
\caption{Maximum Performance gain achieved by the random data dropout using different basis matrices for the fully-connected network depicted in Fig. \ref{fig:fully_connected}.}
\begin{tabular}{|c |c  |c| c |}
\hline
   &  Normalized Hadamard  & Randomly  & Identity     \\
   & matrix &  generated basis & matrix     \\
\hline
 Performance gain  &  2.38  & 2.14  & 2.03     \\
\hline
\end{tabular}
\label{tab:gain_f}
\end{table}

\subsection{Experiment with a CNN}
\begin{figure}[t!]
 \centering
    \includegraphics[width=0.45 \textwidth]{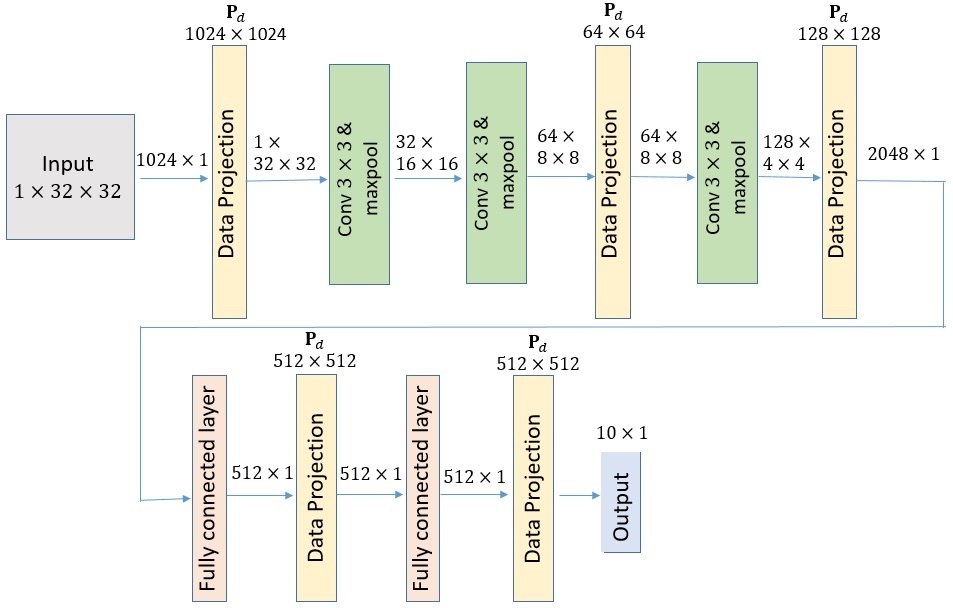}
    \vspace{-0.3cm}
    \caption{The structure of the CNN (during training) used in the second experiment. The network consists of 3 convolutional followed by max-pooling layers plus two fully-connected layers.  }
    \label{fig:CNN_structure}
\end{figure}

\begin{figure}[t!]
 \centering
    \includegraphics[width=0.4 \textwidth]{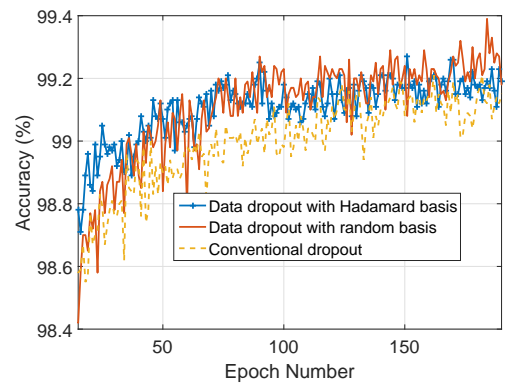}
    \vspace{-0.3cm}
    \caption{Test error versus epoch number for the convolutional network. The network is trained with 15000 images.}
    \label{fig:test_error_conv_c}
\end{figure}

In this experiment, a 5-five layer neural network consisting of 3 convolutional and 2 fully-connected layers is implemented. Each convolutional layer is followed by a max-pooling layer with kernel size and stride equal to 2. Fig. \ref{fig:CNN_structure} illustrates the network structure and the dimension of the data after each layer. The random data dropout is applied to the given data, the output of the second and the third layers, and the output of each fully-connected layer. We did not use data dropout in the output of the first convolutional layer  because the number of extracted feature planes by the first layer is not large. As shown in Fig. \ref{fig:CNN_structure}, first we vectorize the data to apply the data projection. The data projection after each convolutional layer is applied along the channel dimension. Thus, the dimension of the projection matrices after the second and the third convolutional layers which extract 64 and 128 feature planes, respectively, are $64\times 64$ and $128 \times 128$. Similar to the previous experiment, in order to emphasize the overfitting phenomena, we do not train the network with the full training data.  In this experiment, the network is trained with
15000 images sampled randomly from the MNIST training data images. As for the previous experiment, we first train the network for 50 epochs with no data dropout and the saved network is trained for 200 epochs using different basis matrices for the random data dropout. The network parameters are updated using the Adam algorithm \cite{kingma2014adam} with learning rate equal to $10^{-4}$.
Fig. \ref{fig:test_error_conv_c} shows the test error after each epoch. In comparison to the fully-connected network, the performance gain achieved by data dropout is smaller.
In this experiment, the random basis yields the largest performance gain. Table \ref{tab:gain_c} summarizes the  performance gains for different basis matrices.

\begin{table}
\centering
\caption{Maximum Performance gain achieved by the random data dropout using different basis matrices for the convolutional network depicted in Fig. \ref{fig:CNN_structure}}
\begin{tabular}{|c |c  |c| c |}
\hline
   &  Normalized Hadamard  & Randomly  & Identity     \\
   & matrix &  generated basis & matrix     \\
\hline
 Performance gain  &  0.22  & 0.34  & 0.16     \\
\hline
\end{tabular}
\label{tab:gain_c}
\end{table}

\nocite{*}

\bibliographystyle{IEEEtran}
\bibliography{IEEEabrv,bibfile}



\end{document}

%% file: preamble.tex
\usepackage[mathscr]{eucal}
\usepackage[cmex10]{amsmath}
\usepackage{epsfig,epsf,psfrag}
\usepackage{amssymb,amsmath,amsthm,amsfonts,latexsym}
\usepackage{amsmath,graphicx,bm,xcolor,url}
\usepackage[caption=false]{subfig} 
\usepackage{fixltx2e}
\usepackage{array}
\usepackage{verbatim}
\usepackage{bm}
\usepackage{algorithmic, cite}
\usepackage{algorithm}
\usepackage{verbatim}
\usepackage{textcomp}
\usepackage{mathrsfs}
\usepackage{epstopdf}

\catcode`~=11 \def\UrlSpecials{\do\~{\kern -.15em\lower .7ex\hbox{~}\kern .04em}} \catcode`~=13 

\allowdisplaybreaks[3]

\newcommand{\bb}{\mathbf{b}}

\newcommand{\be}{\mathbf{e}}

\newcommand{\bg}{\mathbf{g}}
\newcommand{\bG}{\mathbf{G}}

\newcommand{\bI}{\mathbf{I}}

\newcommand{\bP}{\mathbf{P}}

\newcommand{\bW}{\mathbf{W}}
\newcommand{\bx}{\mathbf{x}}

\DeclareMathAlphabet{\mathbsf}{OT1}{cmss}{bx}{n}
\DeclareMathAlphabet{\mathssf}{OT1}{cmss}{m}{sl}

\DeclareSymbolFont{bsfletters}{OT1}{cmss}{bx}{n}  
\DeclareSymbolFont{ssfletters}{OT1}{cmss}{m}{n}
\DeclareMathSymbol{\bsfGamma}{0}{bsfletters}{'000}
\DeclareMathSymbol{\ssfGamma}{0}{ssfletters}{'000}
\DeclareMathSymbol{\bsfDelta}{0}{bsfletters}{'001}
\DeclareMathSymbol{\ssfDelta}{0}{ssfletters}{'001}
\DeclareMathSymbol{\bsfTheta}{0}{bsfletters}{'002}
\DeclareMathSymbol{\ssfTheta}{0}{ssfletters}{'002}
\DeclareMathSymbol{\bsfLambda}{0}{bsfletters}{'003}
\DeclareMathSymbol{\ssfLambda}{0}{ssfletters}{'003}
\DeclareMathSymbol{\bsfXi}{0}{bsfletters}{'004}
\DeclareMathSymbol{\ssfXi}{0}{ssfletters}{'004}
\DeclareMathSymbol{\bsfPi}{0}{bsfletters}{'005}
\DeclareMathSymbol{\ssfPi}{0}{ssfletters}{'005}
\DeclareMathSymbol{\bsfSigma}{0}{bsfletters}{'006}
\DeclareMathSymbol{\ssfSigma}{0}{ssfletters}{'006}
\DeclareMathSymbol{\bsfUpsilon}{0}{bsfletters}{'007}
\DeclareMathSymbol{\ssfUpsilon}{0}{ssfletters}{'007}
\DeclareMathSymbol{\bsfPhi}{0}{bsfletters}{'010}
\DeclareMathSymbol{\ssfPhi}{0}{ssfletters}{'010}
\DeclareMathSymbol{\bsfPsi}{0}{bsfletters}{'011}
\DeclareMathSymbol{\ssfPsi}{0}{ssfletters}{'011}
\DeclareMathSymbol{\bsfOmega}{0}{bsfletters}{'012}
\DeclareMathSymbol{\ssfOmega}{0}{ssfletters}{'012}

\newtheorem{data model}{Data Model}

\newcommand{\qednew}{\nobreak \ifvmode \relax \else
      \ifdim\lastskip<1.5em \hskip-\lastskip
      \hskip1.5em plus0em minus0.5em \fi \nobreak
      \vrule height0.75em width0.5em depth0.25em\fi}